\documentclass[11pt,a4paper]{article}
\usepackage[utf8]{inputenc}
\usepackage[T1]{fontenc}
\usepackage{lmodern}
\usepackage[margin=2.4cm]{geometry}
\usepackage{graphicx}
\usepackage{booktabs}
\usepackage{array}
\usepackage{multirow}
\usepackage{amsmath,amssymb}
\usepackage{bm}
\usepackage[hidelinks]{hyperref}
\usepackage{caption}
\usepackage{xcolor}
\usepackage{colortbl}

\usepackage{tikz}
\usetikzlibrary{arrows.meta,positioning,decorations.pathreplacing,calc}
\usepackage[section]{placeins}
\definecolor{panelblue}{HTML}{CDD9E5}
\graphicspath{{figures/}}
\usepackage{pdfpages}
\usepackage{fancyhdr}
\fancypagestyle{preprintfirst}{\fancyhf{}\fancyfoot[C]{\parbox{\textwidth}{\centering\small\itshape Preprint, version 2, October 2026.\\[3pt]\normalfont\normalsize\thepage}}}

\title{\bfseries The record is part of the task: matched-record evaluation of text classifiers across maintenance, safety and recall reporting}
\author{Hisham Ihshaish$^{1,2,*}$ \quad Peter Mayhew$^{1,3,*}$ \quad Tasnim M. A. Zayet$^{2}$ \quad Ana Del Amo$^{4}$\\[4pt]
\small $^{1}$School of Computing and Creative Technologies, University of the West of England, Bristol, UK \\
\small $^{2}$Department of Computer Science, Birzeit University, Birzeit, Palestine \\
\small $^{3}$GE Aerospace, Cheltenham, UK \\
\small $^{4}$GE Aerospace, Clearwater, FL, USA}
\date{}
\let\origthebibliography\thebibliography
\renewcommand{\thebibliography}[1]{\origthebibliography{#1}\setlength{\itemsep}{0pt}}
\begin{document}
\maketitle
\thispagestyle{preprintfirst}
{\renewcommand{\thefootnote}{\fnsymbol{footnote}}\footnotetext[1]{Corresponding authors: hisham.ihshaish@uwe.ac.uk; peter.mayhew@geaerospace.com}}

\begin{abstract}\sloppy
\noindent Many operational cases are documented more than once, at different workflow stages and for different purposes, yet model evaluations normally select one of these records before model comparison begins. We treat that selection as part of the evaluation and compare matched records of the same cases under fixed labels and splits in three systems: GE Aerospace repair events, NASA ASRS safety reports and NHTSA vehicle recalls. The design fixes, for any decision a classifier would support, the records that existed and could be read when it was taken, and compares them retrospectively, as written. Across the three GE fields, for events whose label comes from parts transactions independently of the narratives, held-out macro-F1 ranged from 0.33 to 0.91. A difference of 0.46 separated the customer report, written before shop work, from the technician report, written after diagnosis but before the transaction that generates the label. That difference is substantially larger than the representation and architecture differences tested on the same events. The public systems showed different patterns: the NHTSA defect summary remained strongest under every model family tested, whereas the ASRS analyst synopsis outperformed the reporter narrative under learned sequence models but not under lexical baselines. Secondary analyses showed that some model comparisons were also record-dependent. Evaluations should be run on the information available at the intended decision point and should report how both the record and the label were produced.

\end{abstract}

\section{Introduction}\label{sec:intro}

A repair, a safety report or a recall campaign generates several written accounts as it moves through a workflow, written by different people, at different stages, for different purposes. Maintenance work orders in computerised maintenance management systems carry record quality as a standing constraint on downstream analysis~\cite{Brundage2021,Conte2021}. The NASA Aviation Safety Reporting System holds tens of thousands of voluntary reports in its Aircraft category alone, and every vehicle manufacturer must lodge recall filings with its regulator. When such text is classified, evaluation practice selects one of these records, and the selection is rarely treated as part of what is being evaluated. In 1796 the Astronomer Royal, Nevil Maskelyne, dismissed his assistant David Kinnebrook over the times each recorded for the same stellar transits. Bessel later made the episode the founding case of the personal equation~\cite{Schaffer1988}, although Kinnebrook's own letters complicate that account~\cite{Lund2025}. The lesson, that a measurement is read together with the process that produced it, holds for the records that maintenance and safety systems now classify.

We use \emph{record} for any distinctly produced textual account of a case: a named free-text field within one filing, or a separately submitted narrative. The records of one case are not interchangeable accounts of the same content. They differ in who wrote them, what was known when they were written and what they were written to do. A classifier never reads the case; it reads a record of the case. The label it is scored against is produced by a further process, sometimes independent of every record and sometimes inside the workflow that wrote one of them. Measured performance can therefore reflect documentation as well as modelling. The questions such classifications serve are practical ones, which assembly to order before a unit reaches the shop, which reports a safety analyst might examine first and how a regulator might code a campaign that affects an entire fleet. Clinical NLP has met one form of this problem: classifiers confounded by the institution that produced a note~\cite{Ding2023}, and retrieval rankings that fail to carry across documentation settings~\cite{Mikkelsen2026}.

Later records in a workflow can also contain information that did not yet exist when the earlier ones were written. A repair of a cockpit display unit illustrates the point. The customer reports ``Display blanks intermittently in flight''; the technician, after diagnosis, writes ``Fault isolated to processor assembly; key panel checked serviceable''; the repair record states ``Replaced processor assembly''. All three describe one return, and each later record holds information the earlier ones could not. Diagnosis generated new information between the first record and the second, and the completed repair generated more before the third. Because the three records are different information states, each becomes a legitimate model input only once the information it contains exists, and a high score on the third record says nothing about a classifier for the first, which addresses a different problem. In a fleet of more than ten thousand units in service, an evaluation on the wrong record misjudges a triage aid before it is deployed.

Although data source and documentation context are known to condition NLP systems (Section~\ref{sec:related}), much less is known about what happens when alternative records of the same cases are exchanged under a common target.

This study asks two questions on matched records from three record systems, one proprietary and two public (Figures~\ref{fig:workflow} and~\ref{fig:systems}). The systems are GE Aerospace repair events, extending our earlier classification study of those records~\cite{Mayhew2023} with a different question; safety reports from the NASA Aviation Safety Reporting System (ASRS); and vehicle recall campaigns reported to the National Highway Traffic Safety Administration (NHTSA). First, how much does classification performance change when a model reads a different record of the same case? Second, do conclusions about modelling approaches remain the same across those records? Within every comparison, cases, targets and evaluation splits are fixed, and model families and training schedules are held constant (Section~\ref{sec:formulation}). We also explore whether disagreement between records, or between models reading different records of a case, can identify records that warrant review. The practical question behind both is whether an organisation should improve the model, improve the information captured before the decision or change which records the system is allowed to use. Throughout, we distinguish who wrote each record, when in the workflow it was written and for what purpose, and we make the same distinction for the label. The contribution is an evaluation design that places alternative routine records of the same cases on the same held-out scale as the modelling choices tested and ties every comparison to the workflow stage at which the record exists. On one scale, the two differences show which constraint an organisation should address first. That timing also fixes, for any decision the classifier would support, which records an evaluation may use (Equation~\ref{eq:admissible}).

The design is set out in Section~\ref{sec:formulation} and applied to the three systems in Sections~\ref{sec:data} to~\ref{sec:results}. Section~\ref{sec:discussion} draws the engineering implication, that a benchmark is only as valid as the records a system can read when the decision is taken.

\section{Related Work}\label{sec:related}

\subsection{Evaluation validity and data context}\label{sec:rw-validity}
Whether an evaluation supports its claims depends on whether its task and measures represent the phenomenon being claimed. A systematic review of large language model benchmarks finds construct validity to be a recurring concern~\cite{Bean2025construct}. Industrial AI research reaches a similar position from the engineering side: a recent meta-review organises eighty-two data issues across seven stages of the data lifecycle, from source and quality onward, and proposes managing them at every stage~\cite{LiDataIssues2025}. Models can also exploit decision rules that succeed on standard benchmarks yet fail under more demanding conditions~\cite{Geirhos2020}. What counts as leakage depends on the intended use of the model~\cite{Kapoor2023}, and in clinical prediction whether every predictor was available at the moment the model is intended to be used is a standing applicability item~\cite{ProbastAI2025}. The same concern arises here through the record selected as model input.

\subsection{Multiple records and documentation context}\label{sec:rw-provenance}
The most direct evidence that documentation context conditions model behaviour comes from clinical NLP, where classifiers are confounded by the institution that produced a note and provenance-aware adjustment changes their conclusions~\cite{Ding2023}. In clinical embedding-based retrieval, the context variables of the documentation explain as much retrieval-performance variance as the choice of embedding model, and model--corpus interactions make rankings non-portable across documentation settings~\cite{Mikkelsen2026}. Those studies compare institutions, corpora or query conditions. A close same-case design compares note types of the same admissions for a given outcome, with a single model trained on all notes and the note types subsampled to equal length~\cite{Hsu2020notes}, and finds that the most useful note type depends on the outcome. Here the variation is within the case and the procedure is refitted to each record at its full length: the label and the evaluated cases stay fixed while a different routinely produced record of each case is supplied. Multi-view learning starts from the same multiplicity but asks how records should be combined~\cite{Yu2024multiview}, and a systematic review of multi-view document classification finds fusion gains that are consistent but modest~\cite{Mironczuk2026fusion}. Matched-record evaluation first asks what each record does on its own, and in deployed systems a retrieval component may make that choice, which makes it a property of the deployed system as well as of the evaluation. Prediction-time availability, data provenance and documentation context are established concerns~\cite{ProbastAI2025,Ding2023,Mikkelsen2026}, and multi-view classification routinely fits a separate classifier to each view of the same labelled instances~\cite{Mironczuk2019views,Yu2024multiview}. The distinction here is that the views are routine records generated by an organisational workflow, with different producers, purposes and availability times. The performance difference between those records, with the same procedure refitted to each record, is treated as the estimand itself and not as a baseline for fusion. The design then tests whether model comparisons survive exchanging the record while cases, labels and evaluation partitions remain fixed.

\subsection{Maintenance and safety text}\label{sec:rw-quality}
Aviation and maintenance text has been classified with sparse statistical learners~\cite{Tanguy2016,Robinson2015}, recurrent models over embeddings~\cite{Mayhew2023,Nanyonga} and, more recently, domain-pretrained transformers and large language models~\cite{Chandra2023,TikayatRay2023,Kumar2025,Li2026roberta}. Within reliability engineering, maintenance reports and incident narratives now feed the clustering of interventions and the identification of failure mechanisms~\cite{Valcamonico2025,Bhardwaj2025}, models of severity and causal factors for incidents in process safety and power distribution~\cite{Valcamonico2024,KarimiDehkordi2025} and the classification of pilot failures in accident reports~\cite{Macedo2026}. Most evaluations use public safety corpora, mainly ASRS, whose analyst-assigned labels and de-identified text differ in kind from the maintenance records operators hold~\cite{YangHuang2023,Nanyonga2025review}, and work on aviation maintenance records remains comparatively thin~\cite{Kala2022,Scott2024}.

The technical-language-processing literature established maintenance work orders as a distinct NLP object, telegraphic and written under time pressure, with record quality as a persistent constraint~\cite{Brundage2021,MaintNet} that reduces the accuracy of downstream analyses, often invisibly~\cite{Conte2021}. Maintenance text now serves to correct metadata in computerised maintenance management systems (CMMS), retrieve related records, extract causal relations and derive repair subtasks~\cite{Deloose2023,Naqvi2024,Hershowitz2024,Giordano2024}, and maintenance reports are mined for equipment health status and fault diagnosis~\cite{Bhardwaj2023,Chung2023,LiMeijing2025}. Classifier disagreement has identified ambiguous or incorrect CMMS metadata~\cite{Deloose2023}, and the audit here extends that idea to a label generated separately from any narrative. Maintenance research is also moving toward systems that use tools and multiple agents for diagnostic and prescriptive work~\cite{CrespoMarquez2026,Farahani2026}, which raises the question of what information is available to the system at each stage. In the maintenance and safety studies reviewed here, the input record is generally treated as a fixed property of the dataset, chosen before the evaluation begins. The effect of supplying another routine record of the same cases under the same target is not measured.

\section{Matched-record study design}\label{sec:formulation}
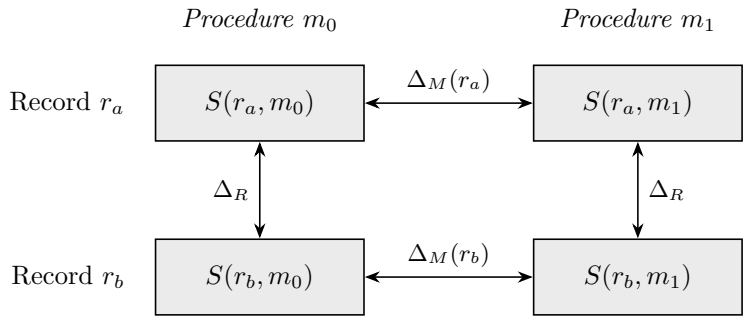
\begin{figure}[!b]\centering
\begin{tikzpicture}[>=Stealth,
  cell/.style={draw, semithick, align=center, minimum height=10mm, text width=25mm, font=\small, fill=black!8},
  arr/.style={<->, semithick}]
\node[font=\small\itshape] at (2.0,1.1) {Procedure $m_0$};
\node[font=\small\itshape] at (7.0,1.1) {Procedure $m_1$};
\node[cell](aa) at (2.0,0) {$S(r_a,m_0)$};
\node[cell](ab) at (7.0,0) {$S(r_a,m_1)$};
\node[cell](ba) at (2.0,-2.3) {$S(r_b,m_0)$};
\node[cell](bb) at (7.0,-2.3) {$S(r_b,m_1)$};
\node[font=\small, anchor=east] at (0.35,0) {Record $r_a$};
\node[font=\small, anchor=east] at (0.35,-2.3) {Record $r_b$};
\draw[arr](aa.south)--(ba.north) node[midway,left,font=\footnotesize]{$\Delta_R$};
\draw[arr](ab.south)--(bb.north) node[midway,right,font=\footnotesize]{$\Delta_R$};
\draw[arr](aa.east)--(ab.west) node[midway,above,font=\footnotesize]{$\Delta_M(r_a)$};
\draw[arr](ba.east)--(bb.west) node[midway,above,font=\footnotesize]{$\Delta_M(r_b)$};
\node[font=\footnotesize\itshape, align=center] at (4.5,-3.55) {$D=\Delta_M(r_a)-\Delta_M(r_b)$: does the model comparison change with the record?};
\end{tikzpicture}
\caption{The matched-record design as a record-by-procedure table of held-out scores $S(r,m)$ on the same cases and labels: $\Delta_R$ compares records under one refitted procedure, $\Delta_M$ compares procedures on one record, and $D$ asks whether the procedure comparison changes with the record.}
\label{fig:workflow}
\end{figure}

Each system supplies matched records: case $i$ carries records $X_i^{(1)},\dots,X_i^{(V)}$ and one label $Y_i$. We fix a modelling procedure $m$, an architecture with its preprocessing and training schedule, and write $S(r,m)$ for the held-out macro-F1 (the F1 score averaged with equal weight over classes) of the model that procedure fits on record $r$, scored over the shared held-out cases, cases set aside from all training (Figure~\ref{fig:workflow}). Two estimands are defined on this table of scores. The record contrast
\begin{equation}
\Delta_R(r_a,r_b;m)\;=\;S(r_a,m)-S(r_b,m)
\label{eq:delta}
\end{equation}
compares two records under one procedure: cases, targets, splits, model family and training schedule are fixed, and each record's model is fitted and evaluated separately. $\Delta_R$ is therefore the end-to-end consequence of changing the record, with the procedure refitted to each record. The model contrast $\Delta_M(r;m_1,m_0)=S(r,m_1)-S(r,m_0)$ compares two models on one record, and its stability across records is measured by the interaction
\begin{equation}
D\;=\;\Delta_M(r_a;m_1,m_0)-\Delta_M(r_b;m_1,m_0),
\label{eq:interaction}
\end{equation}
which is nonzero when the observed benefit of a modelling choice depends on the record read. When the two model contrasts differ in sign, the ranking of the two models reverses between records. We define one further quantity, used once: the cross-record transfer $T(r_a\!\to\! r_b;m)$, the score of the model fitted on record $r_a$ when it reads record $r_b$ of the same cases, which combines record differences with the shift between training and evaluation text (Section~\ref{sec:transfer}). Every model-performance comparison in the study is one of these quantities, with paired inference over the shared test cases (Section~\ref{sec:protocol}).

\paragraph{Decision-time availability:}
Let $H_i(\tau)$ denote the information about case $i$ that exists by workflow time $\tau$, and let record $v$ be produced at time $\tau_v$ by a recording process $g_v$, its author, purpose, selection and wording, so that $X_i^{(v)}=g_v\bigl(H_i(\tau_v)\bigr)$. Two records of one case can differ because $\tau_a\neq\tau_b$, so that different information existed, or because $g_a\neq g_b$, so that the same information was recorded differently, and $\Delta_R$ measures the joint consequence of both without separating them. A decision taken at time $\tau$ can read only records that exist by then and that the deployed system is permitted to access. Writing $\mathcal P$ for the set of records the system may access, the admissible set is
\begin{equation}
\mathcal A_i(\tau)=\{\,v:\tau_v\le\tau,\ v\in\mathcal P\,\}.
\label{eq:admissible}
\end{equation}
An evaluation is decision-time valid with respect to information availability only if every input it uses lies in $\mathcal A_i(\tau)$, whether the input is fixed in advance or selected at run time by a retrieval policy, $\pi(i,\tau)\in\mathcal A_i(\tau)$ for a single record and $\pi(i,\tau)\subseteq\mathcal A_i(\tau)$ for several.

The three systems place the design under three of the possible relationships between record production and label production (Figure~\ref{fig:systems}). The GE label is produced independently of every narrative, from parts transactions; the NHTSA label is the regulator's coding of the report that contains the fields; the ASRS label is assigned inside the analyst workflow that also writes one of the records. The systems differ accordingly in how directly a measured contrast can be interpreted: the record comparison is most directly interpretable where the label is produced independently of every record, and most entangled where record and label are co-produced. Independence bears on how the comparison is interpreted, and the validity of the label is a separate question. The transaction-derived label has failure modes of its own (Section~\ref{sec:error}).

\paragraph{Identification and scope:}
Each record contrast is estimated within one system, on the same held-out cases, label and partition, with the procedure refitted to each record, so differences between the three systems enter no primary comparison. Within a system, the analyses that mask vocabulary, share a representation, match length, group duplicates and split by filing time test named alternative explanations for a difference. None of them decomposes $\Delta_R$ into the author, stage, purpose, length and wording that change together with the record. The comparison is retrospective, reading each record as it was written, so it measures the consequence of the record choice under current capture practice and leaves the effect of changing that practice to a prospective study.

\section{Data}\label{sec:data}

\begin{figure}[tbp]\centering
\resizebox{\textwidth}{!}{%
\begin{tikzpicture}[>=Stealth,
  case/.style={draw, semithick, align=center, minimum height=11mm, text width=21mm, font=\footnotesize},
  rec/.style={draw, semithick, align=center, text width=35mm, font=\footnotesize, fill=black!8, inner sep=2.5pt},
  prod/.style={draw, semithick, align=center, minimum height=10mm, text width=23mm, font=\footnotesize},
  tgt/.style={draw, very thick, align=center, minimum height=10mm, text width=25mm, font=\footnotesize},
  plab/.style={font=\bfseries\small},
  arr/.style={->, semithick}]
\node[plab] at (-1.35,1.35) {(a)};
\node[case](gcase) at (0,0) {{\bfseries GE Aerospace}\\repair event};
\node[rec](gr1) at (5.4,1.15) {Customer-reported fault\\{\footnotesize\itshape operator, pre-shop}};
\node[rec](gr2) at (5.4,0)    {Technician report\\{\footnotesize\itshape technician, diagnosis}};
\node[rec](gr3) at (5.4,-1.15){Repair action\\{\footnotesize\itshape technician, post-repair}};
\node[prod](gpt) at (10.4,-2.0) {Parts\\transactions};
\node[tgt](gtg) at (14.4,-2.0) {Outcome label\\{\footnotesize 4 classes}};
\draw[arr](gcase.east)--(gr1.west); \draw[arr](gcase.east)--(gr2.west); \draw[arr](gcase.east)--(gr3.west);
\draw[arr](gcase.south) |- (gpt.west);
\draw[arr](gpt.east)--(gtg.west);
\node[font=\footnotesize\itshape, color=black!60] at (4.4,-2.45) {from the repair itself, independent of the narratives};
\draw[black!20, thin] (-1.4,-3.0)--(15.9,-3.0);
\node[plab] at (-1.35,-3.35) {(b)};
\node[case](acase) at (0,-4.7) {{\bfseries NASA ASRS}\\safety report};
\node[rec](ar1) at (5.4,-4.15) {Reporter narrative\\{\footnotesize\itshape reporter, at submission}};
\node[rec](ar2) at (5.4,-5.3){Supplemental narrative\\{\footnotesize\itshape second reporter}};
\node[prod](aan) at (10.4,-4.7) {ASRS analyst processing};
\node[rec, text width=25mm](asy) at (14.4,-4.05) {Analyst synopsis\\{\footnotesize\itshape analyst, same step}};
\node[tgt](atg) at (14.4,-5.4) {Primary problem\\{\footnotesize binary task}};
\draw[arr](acase.east)--(ar1.west); \draw[arr](acase.east)--(ar2.west);
\draw[arr](ar1.east)--(aan.west); \draw[arr](ar2.east)--(aan.west);
\draw[arr](aan.east)--(asy.west); \draw[arr](aan.east)--(atg.west);
\node[font=\footnotesize\itshape, color=black!60](aoi) at (10.4,-3.5) {other analyst information};
\draw[arr, dashed, black!60](aoi.south)--(aan.north);
\draw[black!20, thin] (-1.4,-6.7)--(15.9,-6.7);
\node[plab] at (-1.35,-7.05) {(c)};
\node[case](ncase) at (0,-8.55) {{\bfseries NHTSA}\\recall campaign};
\node[rec](nr1) at (5.4,-7.4) {Defect summary\\{\footnotesize\itshape manufacturer report}};
\node[rec](nr2) at (5.4,-8.55){Consequence\\{\footnotesize\itshape manufacturer report}};
\node[rec](nr3) at (5.4,-9.7) {Remedy\\{\footnotesize\itshape manufacturer report}};
\node[prod](nan) at (10.4,-8.55) {NHTSA analysis of the report};
\node[tgt](ntg) at (14.4,-8.55) {Component class\\{\footnotesize 16 classes}};
\draw[arr](ncase.east)--(nr1.west); \draw[arr](ncase.east)--(nr2.west); \draw[arr](ncase.east)--(nr3.west);
\draw[arr](nr1.east)--(nan.west); \draw[arr](nr2.east)--(nan.west); \draw[arr](nr3.east)--(nan.west);
\draw[arr](nan.east)--(ntg.west);
\node[font=\footnotesize\itshape, color=black!60](nfr) at (10.4,-7.35) {full Part 573 report};
\draw[arr, dashed, black!60](nfr.south)--(nan.north);
\end{tikzpicture}}
\caption{The three record systems on one plan: each case yields several records (shaded, producer and stage inside) and one target (thick border). The GE target comes from parts transactions, bypassing the narratives; the ASRS analyst and the NHTSA regulator code their targets from the reports, and the ASRS analyst also writes the synopsis. Dashed inputs mark information beyond the modelled records.}
\label{fig:systems}
\end{figure}
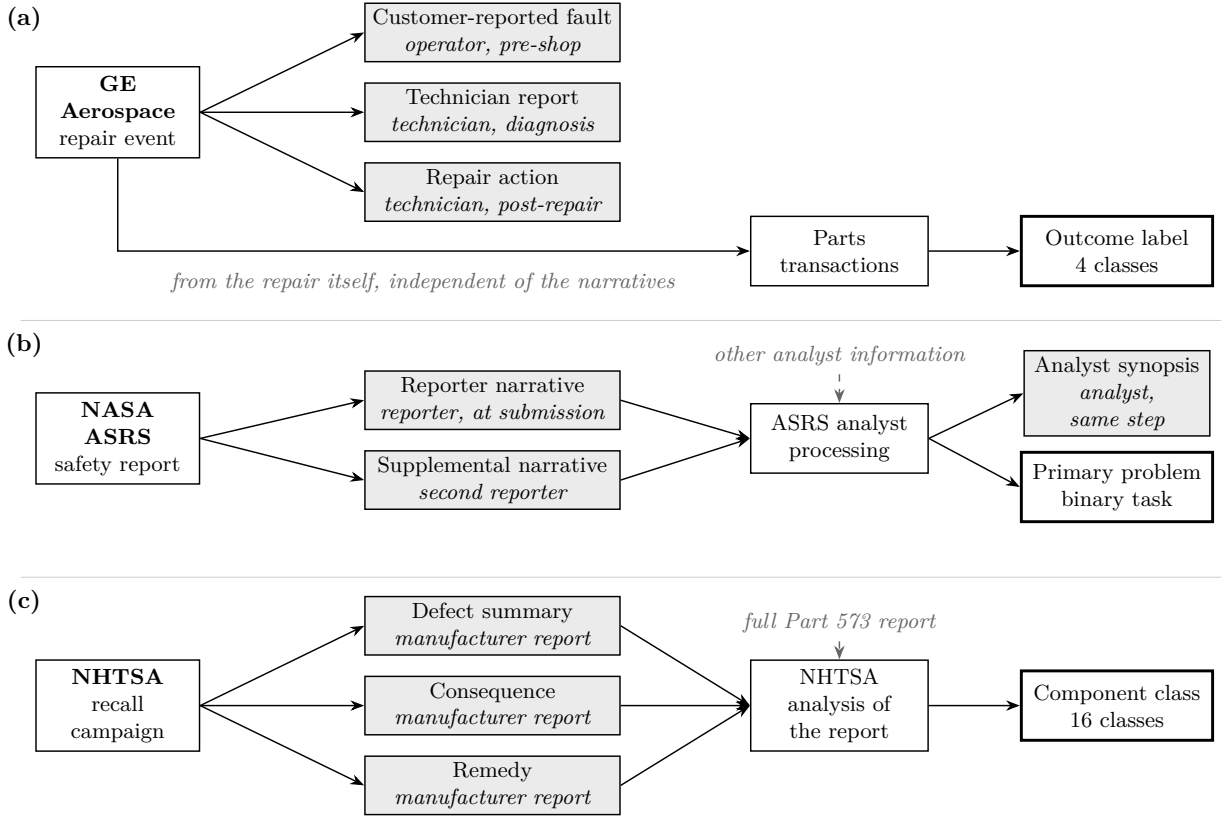

Three record systems are used, each supplying three records per case (Figure~\ref{fig:systems}; Table~\ref{tab:provenance}). Held-out partitions are constructed as described in Section~\ref{sec:protocol}, with the exact-duplicate rule given with the NHTSA dataset, and Figure~\ref{fig:lengths} shows the token-length distributions of the public systems' records. Tables, figures and sections numbered S1 onward are in the Supplementary Material.

\begin{table}[tbp]\centering
\caption{The records and their provenance, with median lengths in tokens. GE medians cannot be exported (dashes; mean character lengths in Section~S1). Each record's relation to the label is shown in Figure~\ref{fig:systems}.}\label{tab:provenance}
\footnotesize\setlength{\tabcolsep}{5pt}
\begin{tabular}{@{}>{\raggedright\arraybackslash}p{4.6cm}>{\raggedright\arraybackslash}p{4.6cm}rr@{}}
\toprule
Record & Producer; stage & Median & Records \\
\midrule
\addlinespace[3pt]
\rowcolor{black!7}\multicolumn{4}{@{}l}{\textbf{GE Aerospace repair events}\quad\emph{target: replaced assemblies, from parts transactions (4 classes)}}\\
\addlinespace[3pt]
Customer-reported fault & operator; pre-shop & -- & 5792 \\
Technician report & technician; diagnosis & -- & 5792 \\
Repair action & technician; post-repair & -- & 5792 \\
\addlinespace[6pt]
\rowcolor{black!7}\multicolumn{4}{@{}l}{\textbf{NASA ASRS safety reports (Aircraft task)}\quad\emph{target: analyst-assigned primary problem (binary)}}\\
\addlinespace[3pt]
Reporter narrative & reporter; submission & 178 & 89194 \\
Supplemental narrative & second reporter; submission & 91 & 7637 \\
Analyst synopsis & analyst; processing & 19 & 89186 \\
\addlinespace[6pt]
\rowcolor{black!7}\multicolumn{4}{@{}l}{\textbf{NHTSA recall campaigns}\quad\emph{target: NHTSA component classification of the campaign (16 classes)}}\\
\addlinespace[3pt]
Defect summary & manufacturer report & 41 & 16626 \\
Consequence & manufacturer report & 19 & 16626 \\
Remedy & manufacturer report & 35 & 16626 \\
\addlinespace[2pt]
\bottomrule
\end{tabular}
\end{table}

\subsection{GE Aerospace LRU dataset}
Each time an avionics line-replaceable unit (LRU) is removed from an aircraft and returned to a maintenance, repair and overhaul (MRO) shop, the repair generates free text at successive workflow stages, and these narratives feed failure reporting and corrective-action processes~\cite{Pelt2019,ThurstonB2018,Candell2009}. The dataset covers one civil cockpit-display LRU family with more than 10k units in service: 5792 repair events after the exclusions needed for reliable transaction linkage and unambiguous classes (Section~S1), of which 4633 train and 1159 are held out. Three free-text fields of each event are modelled, the \emph{customer-reported fault}, the \emph{technician report} and the \emph{repair action} (Figure~\ref{fig:systems}), of the form shown by the synthetic record in Section~\ref{sec:intro}.  The label is the assembly actually replaced, taken from the parts transactions and joined to the narratives by the return identifier. Replaced parts are the operational reference because they record the maintenance action independently of the narratives and drive spares forecasting. Their agreement with the narratives is examined in Section~\ref{sec:consequences}. The task covers the two most frequently replaced sub-assemblies as four mutually exclusive classes, chosen for label determinacy as much as frequency.

The fields differ in when they exist relative to the label (Figure~\ref{fig:systems}). The customer text exists before any shop work and the technician report is written after diagnosis has identified the fault, so both precede the parts transaction that produces the label, at different information stages. The repair action is written during or after the repair that generates the label, so classifying it is a post-outcome task, automated coding and narrative--transaction reconciliation, in which the field's account of the outcome is legitimately available. How much of that field's value is lexical restatement of the outcome is tested by masking in Section~\ref{sec:oov}.

\subsection{NASA ASRS dataset}\label{sec:nasadata}
ASRS holds voluntary, de-identified safety reports, each with one \emph{primary problem} assigned by an ASRS analyst among 18 categories of markedly unequal prevalence. We reformulate classification as independent binary tasks with majority-class undersampling to 50:50 and report the largest equipment-related category in full: all 44597 Aircraft records through 2021 with a non-blank primary problem, against an equal undersample of the remaining categories.

Two further records of each case accompany the reporter's narrative (Table~\ref{tab:provenance}). The \emph{synopsis} is a short summary written by the analyst who processes the report and assigns the primary problem, drawing on overflow text and fields absent from the public record~\cite{ASRSaboutdata}. On the Aircraft task it accompanies 89186 of 89194 records, at a median 19 tokens against the narrative's 178. Where two parties reported the same event, the second reporter's \emph{supplemental narrative} is attached, and 7637 task records carry one, at a median 91 tokens. Analysts merge multiple reports of one event into one database record, choose which reports receive detailed analysis and edit narratives during processing and de-identification, which replaces tail numbers, operators and equipment identifiers with placeholders~\cite{ASRSprogram}.

\begin{figure}[!b]\centering
\includegraphics[width=\textwidth]{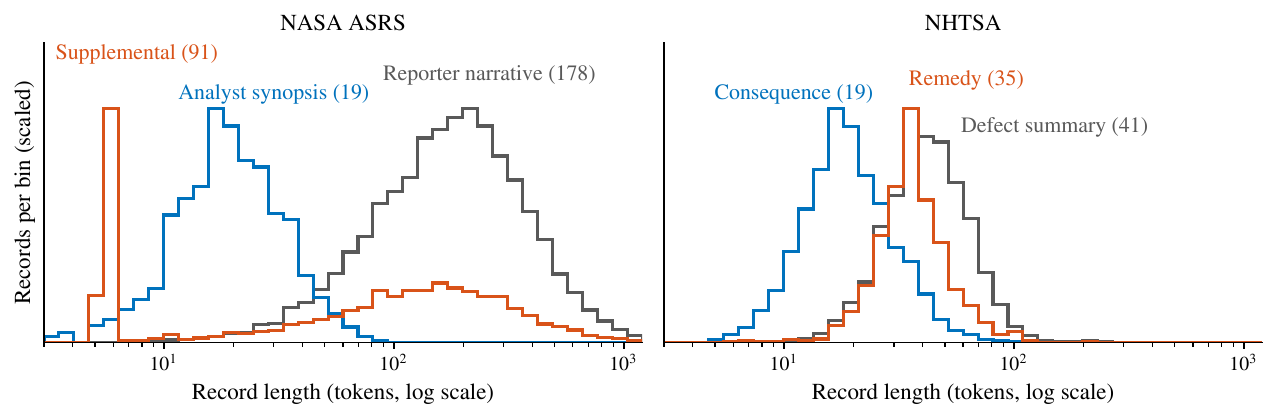}
\caption{Token-length distributions of the records of the same cases in the two public systems, as log-binned histograms scaled to their highest bin with medians in parentheses. The ASRS records of one case differ in length by an order of magnitude and the supplemental narrative is bimodal, with a short-record mode and a full-account mode, whereas the NHTSA fields are similar in length. GE distributions cannot be exported (field statistics in Section~S1).}
\label{fig:lengths}
\end{figure}

\subsection{NHTSA recall campaigns}\label{sec:nhtsadata}
Every United States vehicle recall campaign is reported to NHTSA under 49 CFR Part 573, and the recall record carries three narrative fields from that report: a \emph{defect summary} describing the problem, a \emph{consequence} field describing the safety risk and a \emph{remedy} field describing the corrective action. The target is NHTSA's component classification of the campaign, collapsed to the top level of its taxonomy. NHTSA derives it from its own analysis of the manufacturer's report, with acknowledged subjectivity in the mapping~\cite{NHTSAcompletion}, so the target is a coding of the case by the regulator, separate from the text fields. All campaigns with 2000--2026 campaign numbers were retrieved through the public recall API~\cite{NHTSAapi}, 16626 after de-duplication, and categories with at least 300 campaigns define a 16-class task with 3325 campaigns held out. Boilerplate recurs across filings, 39\% of consequence texts being exact duplicates of another campaign's, so campaigns sharing an identical text in any field are confined to one side of the split. The fields are short (medians 41, 19 and 35 tokens) and the component name can appear verbatim in any of them, which motivates the class-vocabulary masking condition. Campaign numbers encode the filing year, which supports the out-of-time check in Section~\ref{sec:nhtsa} (class list and snapshot details in Section~S1).

\section{Models and analysis}\label{sec:methods}

We use four kinds of model, each for a different diagnostic purpose, with shared-representation controls that separate the record from representation choice where a comparison requires it (Section~\ref{sec:models}). Preprocessing is limited to lower-casing and punctuation stripping before tokenisation into whitespace-delimited alphanumeric units, keeping part identifiers and shorthand intact, with no stemming, lemmatisation or stop-word removal.

The task is single-label classification of a tokenised narrative, with $C=4$ assembly classes for GE, $C=2$ per NASA task and $C=16$ component classes for NHTSA. Each narrative is read through a frozen word embedding, a learned mapping from words to vectors that stays fixed while the classifier trains. Tokens outside an embedding's vocabulary map to a single shared fallback vector. Representation-specific differences, including vocabulary coverage and out-of-vocabulary handling, are analysed in Section~S5.

\subsection{Models}\label{sec:models}\label{sec:representations}\label{sec:classifiers}
The lexical baseline is word TF-IDF (term frequency weighted by inverse document frequency) with linear classifiers, plus a character n-gram variant, and asks whether surface vocabulary alone carries a difference. The learned sequence model is a bidirectional long short-term memory network (BiLSTM)~\cite{Hochreiter1997,Schuster1997}, which reads the record as an ordered sequence over frozen word embeddings. Its embeddings are word2vec vectors trained on the corresponding training record or field. The general-purpose GloVe-200~\cite{Pennington2014a} serves as a shared-representation control, and on GE the comparison also uses Avi2Vec, a word2vec embedding trained on aviation maintenance text~\cite{Mayhew2023,Mikolov2013,Chiu2016}. The pooling comparator is a perceptron over the same frozen embeddings with mean pooling, which averages the word vectors and discards word order. It separates sequence reading from representation and supplies $m_0$ in Equation~\ref{eq:interaction}. The pretrained transformer, RoBERTa-base~\cite{Liu2019roberta}, is fine-tuned end-to-end on each public task and asks whether the results persist under a pretrained contextual encoder. No pretrained checkpoint can be brought into the GE computing environment, so its transformer comparator is trained from scratch. Further controls and every architectural and training constant are in Sections~S2 and~S3. Together these models test whether the record differences persist across lexical, sequential and contextual representations.

\subsection{Inference}\label{sec:protocol}
We split each dataset once into 80\% training and 20\% held-out test data before any preprocessing or tuning, fitting tokeniser and vocabulary statistics on the training partition only. Performance is summarised by macro-averaged F1, weighting the operationally distinct classes equally. Under the class imbalance of the GE task, accuracy would reward majority-class prediction. Each neural configuration is estimated by three independent trainings of the same procedure. A difference is reported as its mean over trainings, written $\bar\Delta$ in Table~\ref{tab:estimands}, with a paired bootstrap 95\% interval obtained by resampling the shared held-out cases jointly, conditional on the trained models. The range over trainings is given separately. The training-averaged estimand and the resample and permutation counts per family are in Section~S2. Significance statements come from paired approximate-randomisation tests, Holm-corrected within each preplanned family of comparisons. Stratified 10-fold cross-validation appears in the supplement as description only~\cite{Nadeau2003}. The paper's conclusions rest on the held-out paired estimates. Unless a model is named, matched-record differences are reported for the record- or field-trained word2vec BiLSTM, and for the Avi2Vec BiLSTM on GE.

\subsection{Comparator implementation and sensitivity analyses}\label{sec:reproprotocol}\label{sec:viewsproto}
We built an independent implementation of the GE and ASRS pipelines to test dependence on the initial codebase, and the two agree (Section~S3). The principal analyses carry an evidential status, preplanned, prospective extension, post hoc or prior-study design, defined and listed in Section~S7. Within each record the sequence and pooling models share that record's embedding, so each $\Delta_M$ in Equation~\ref{eq:interaction} is representation-matched.

\subsection{Audit procedure for model--label disagreements}\label{sec:auditproto}
Misclassified test records are manually reviewed against their narratives and coded as model errors, narrative--transaction inconsistencies or faults outside the label space (category definitions and worked examples in Section~S6). Coders saw narrative, transaction label and prediction together, so anchoring toward the model's reading cannot be excluded. No formal inter-rater statistic was recorded and correctly classified records were not audited. The audit covers all 168 technician-input misclassifications in the initial extraction's 1279-record test set and predates the removal of the earlier-build-variant records. Of the audited records, 162 remain in the final dataset (Section~S6). The audit was performed by GE maintenance engineers and is treated as exploratory.

\section{Results}\label{sec:results}

Table~\ref{tab:estimands} collects the matched-record contrasts and secondary quantities with paired uncertainty. All inferential results use the held-out comparisons (Section~\ref{sec:protocol}).

\begin{table}[!b]\centering
\caption{Held-out record and model comparisons. Panel A gives matched-record differences $\bar\Delta$ with the same procedure fitted separately to each record. Panel B gives the cross-record transfer $T$ and the secondary quantities. Values are three-training means with paired case-bootstrap 95\% intervals and the range over trainings. *GE interval endpoints could not be exported. Status vocabulary is defined in Section~S7.}\label{tab:estimands}
\small\setlength{\tabcolsep}{3pt}
\begin{tabular}{llcc>{\raggedright\arraybackslash}p{2.0cm}}
\toprule
System & Contrast & $\bar\Delta$ [95\% CI] & Range over trainings & Status \\
\midrule
\addlinespace[2pt]
\rowcolor{black!7}\multicolumn{5}{@{}l}{\textbf{A: Matched records (fitted separately)}}\\
\addlinespace[2pt]
GE & customer $\rightarrow$ technician & $+0.456$* & $+0.419$ to $+0.492$ & prior-study design \\
GE & technician $\rightarrow$ repair action & $+0.127$* & $+0.104$ to $+0.152$ & prior-study design \\
ASRS & synopsis $-$ narrative & $+0.015$ [0.011, 0.019] & $+0.011$ to $+0.021$ & preplanned \\
ASRS & synopsis $-$ narrative, RoBERTa & $+0.015$ [0.011, 0.019] & $+0.011$ to $+0.018$ & prospective extension \\
NHTSA & summary $-$ consequence & $+0.115$ [0.099, 0.131] & $+0.113$ to $+0.118$ & preplanned \\
NHTSA & summary $-$ remedy & $+0.148$ [0.131, 0.165] & $+0.129$ to $+0.160$ & preplanned \\
\addlinespace[4pt]
\rowcolor{black!7}\multicolumn{5}{@{}l}{\textbf{B: Cross-record transfer and secondary quantities}}\\
\addlinespace[2pt]
ASRS & primary $\to$ supplemental ($T$) & $+0.179$ [0.157, 0.202] & $+0.170$ to $+0.187$ & preplanned \\
ASRS & comparable-length subset & $+0.010$ [$-$0.013, 0.033] & $+0.001$ to $+0.017$ & post hoc \\
ASRS & interaction $D$ (Eq.~\ref{eq:interaction}) & $+0.017$ [0.013, 0.021] & $+0.014$ to $+0.020$ & post hoc \\
\bottomrule
\end{tabular}
\end{table}

\subsection{Magnitude of the record differences}\label{sec:gradient}\label{sec:views}\label{sec:repro}
\begin{figure}[b]\centering
\includegraphics[width=0.72\textwidth]{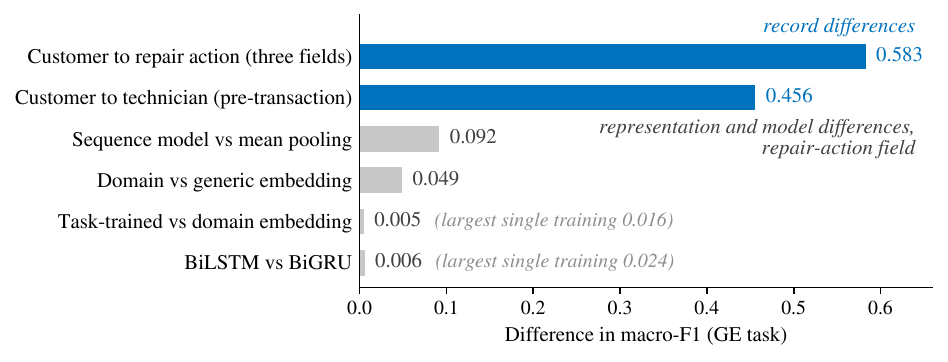}
\caption{Record differences against representation and model differences on one scale, GE task, held-out evaluation. All entries are means over three trainings (Table~S5); the bracketed values on the two smallest rows are the largest single-training differences.}
\label{fig:magnitude}
\end{figure}

Under the Avi2Vec--BiLSTM procedure the three GE fields reach macro-F1 of 0.327 on the held-out cases (customer-reported fault), 0.783 (technician report) and 0.910 (repair action), means over three independent trainings. Reading a different field of the same events therefore moves performance by 0.583 across the three fields, and by 0.456 between the customer and technician fields, both written before the transaction that generates the label. Customer to technician is $+0.419$ to $+0.492$ in every training, technician to repair action $+0.104$ to $+0.152$. The same ordering appears under the TF-IDF baselines and both recurrent architectures. The gain appears in every class and is largest for the rarest (the rarest class's values in Section~S3). Figure~\ref{fig:magnitude} places these record differences against the representation and model differences measured on the same events under the same evaluation: the largest modelling difference tested moves performance by 0.092, the record by up to 0.583. The quantities in Figure~\ref{fig:magnitude} are descriptive magnitudes and are not components of a variance decomposition.

The ASRS analyst's synopsis outperforms the reporter's narrative in every training of a learned sequence model, by 0.011--0.021 macro-F1 across both embeddings, averaging $+0.015$ [0.011, 0.019]. The advantage is absent under the lexical baselines: word TF-IDF is level across the records, and character n-grams favour the narrative nominally (per-configuration results in Table~S9). The fine-tuned RoBERTa shows the same advantage, averaging $+0.015$ [0.011, 0.019], an estimate that coincides with the recurrent models' to three decimals. The primary--supplemental comparison produces the largest raw ASRS difference, and it is a cross-record transfer: fixed narrative models reading the second reporter's account of the same 1530 held-out cases lose 0.165--0.187 macro-F1 in all six models. The number combines record differences with the shift between training and evaluation text, and Section~\ref{sec:transfer} separates the two.

NHTSA component classification is strongest from the defect summary, lower from the consequence field and lowest from remedy text under both TF-IDF and the field-trained BiLSTM (values per run in Section~S4). The summary exceeds the consequence field by $+0.115$ [0.099, 0.131] and the remedy field by $+0.148$ [0.131, 0.165], with 11 of 12 paired contrasts Holm-significant. Under the BiLSTM the per-training differences span $+0.113$ to $+0.118$ and $+0.129$ to $+0.160$ (Table~\ref{tab:estimands}). TF-IDF, a single deterministic evaluation, gives $+0.094$ and $+0.143$. Chosen as different record and label production settings, the two public systems test whether record sensitivity appears there and are not replications of the GE task.

\subsection{Interpreting the record differences}\label{sec:transfer}\label{sec:nhtsa}\label{sec:representation}\label{sec:oov}\label{sec:transformer}
The GE ordering follows the workflow stage at which each field is written (Section~\ref{sec:data}) and is already visible to lexical models, where TF-IDF rises from 0.362 on customer text to 0.891 on repair actions. Explicit outcome terminology contributes to the later fields without accounting for the difference, since a keyword rule over the curated outcome terms reaches 0.292 on the repair-action field and masking those terms lowers the BiLSTM by 0.013 with GloVe-200 and 0.038 with Avi2Vec. Representation choice is smaller again: Avi2Vec exceeds GloVe-200 by 0.049 on repair-action text, 0.028 of it associated with tokens that only the domain vocabulary covers, and the advantage was not observed on the de-identified ASRS text (Sections~S3 and~S5).

The ASRS synopsis result is smaller and harder to interpret, because synopsis and target are produced within the same analyst workflow (Section~\ref{sec:formulation}). Literal category-label tokens alone do not explain the advantage: masking every token that appears in the category labels, in both records, costs the synopsis 0.005 and leaves it above the narrative, and the record with fewer label tokens is the one that scores higher. The direction held across the five redrawn balanced tasks. The much larger transfer loss on the supplemental account behaves differently, falling to about zero as the second account approaches the primary in length (Figure~\ref{fig:asrsviews}, right). With the selection and editing documented in Section~\ref{sec:nasadata}, that pattern points to completeness and curation, and the analyses do not support reporter identity alone as an explanation (all conditions in Section~S4).

\begin{figure}[!b]\centering
\includegraphics[width=\textwidth]{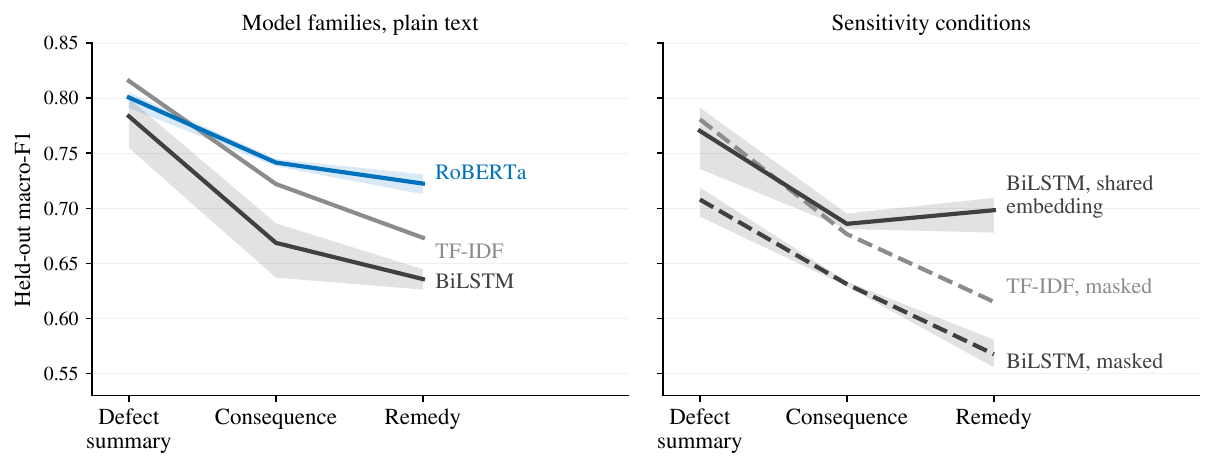}
\caption{Held-out macro-F1 across the three NHTSA fields under the plain model families (left) and under the sensitivity conditions (right), with the class-label vocabulary masked for TF-IDF and the BiLSTM and one embedding shared across the fields for the BiLSTM. Masking lowers absolute performance while the defect summary remains strongest, and the shared embedding makes the consequence--remedy ordering less stable.}
\label{fig:nhtsafields}
\end{figure}

The NHTSA ordering follows the purpose of each field and holds under three different conditions. The fields are similar in length (Figure~\ref{fig:lengths}), so length does not explain it. Masking every token that occurs in the sixteen class labels, at training and evaluation, lowers every score yet preserves the full ordering, all twelve masked contrasts Holm-significant. Trained on campaigns filed 2000--2021 and evaluated on the 2022--2026 filings, the summary remains strongest under both models in every training, all six prospectively specified BiLSTM contrasts Holm-significant. Grouping near-duplicate filings to one side of the split, token-set Jaccard $\geq$0.80 in any field, affected 39.5\% of the held-out campaigns in the main split. It changes absolute scores, which are not comparable across the two test samples, and leaves the ordering intact in every training (prospective extension). The two lower fields are less separable: with one embedding shared across the fields the consequence--remedy difference largely disappears while the summary stays best (Figure~\ref{fig:nhtsafields}, right). The difference therefore extends beyond literal component-name overlap and is associated with where the content relevant to the target is expressed within the filing (Section~S4 and Figure~S3). Table~\ref{tab:reporting} asks an evaluation to report who wrote the record, when and for what purpose, the three properties that organise the differences across the systems.

\subsection{Secondary: record dependence of model comparisons}\label{sec:modelview}
\begin{figure}[!b]\centering
\includegraphics[width=\textwidth]{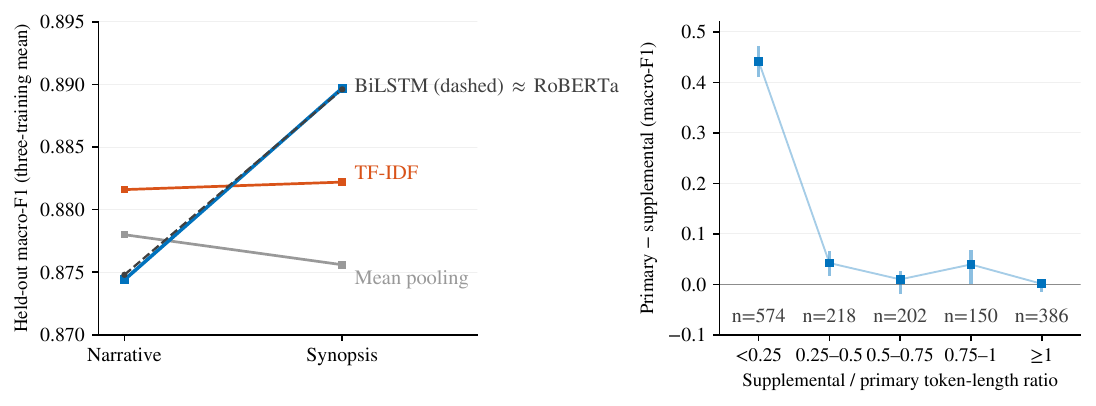}
\caption{ASRS record dependence appears differently across model families and across record completeness. On the left, held-out macro-F1 of each model family on the narrative and on the synopsis, as three-training means on a y-axis spanning 0.870--0.895, where the BiLSTM (dashed) and RoBERTa lie within 0.001 of each other, near 0.875 on the narrative and 0.890 on the synopsis. On the right, the cross-record transfer loss on the supplemental narrative by the ratio of supplemental to primary length, as mean and range over six trainings with bin sizes shown.}
\label{fig:asrsviews}
\end{figure}

Except for the RoBERTa comparisons, a prospective extension, these secondary analyses are post hoc. On the same ASRS cases the ranking of the lexical baseline against the sequence model reverses between the records in every training pair, TF-IDF exceeding the BiLSTM on narratives and trailing it on synopses (Figure~\ref{fig:asrsviews}, left). The benefit of sequence reading over pooling is also larger on synopses. The interaction $D$ of Equation~\ref{eq:interaction} averages $+0.017$ [0.013, 0.021] under the record-trained embeddings and persists at reduced size, though less uniformly, under the shared-representation control. RoBERTa reproduces the synopsis advantage at the same size as the recurrent models, and its benefit over the BiLSTM does not differ by record ($D$ spans $-0.006$ to $+0.006$), while its benefit over TF-IDF does ($+0.011$ to $+0.018$, every interval excluding zero). The record dependence therefore lies between lexical and learned sequence models, and the pretrained encoder and the recurrent reader did not differ by record (per-configuration values and the NHTSA RoBERTa results in Section~S4).

\subsection{Exploratory: record disagreement as review evidence}\label{sec:consequences}\label{sec:error}
The remaining analyses are exploratory. Combining the two ASRS records by mean probability exceeds the better single record by 0.005--0.013 in every configuration (all $p_{\mathrm{Holm}}\leq0.0104$), a gain of the size the multi-view literature reports (Section~\ref{sec:rw-provenance}). Cross-record disagreement, which affects 11.3\% of held-out cases, captured more of the reference model's errors than selecting the same number of least confident predictions at every review budget from 1\% to 30\%. Given the balanced task and the synopsis's origin (Section~\ref{sec:nasadata}), this supports prioritising post-coding review and does not estimate screening yield in service (Section~S6).

The GE audit turns to disagreement between the model's reading of the narrative and the transaction-derived label. Of the 168 technician-text misclassifications audited in the initial extraction (Section~\ref{sec:auditproto}), three quarters (126) were coded as narrative--transaction inconsistency, a fifth (35) as model errors and seven as faults outside the label space. The headline share lies between 74.1\% and 77.8\% under any coding of the six audited records later removed from the dataset (Section~S6).
Within the limits of Section~\ref{sec:auditproto}, narrative--transaction inconsistency accounts for most of the apparent model error among the contested records.

\section{Discussion}\label{sec:discussion}

The 0.456 gap between the GE customer and technician fields opens before the parts transaction exists, and every model change we tested moved the score by less, 0.049 for the domain embedding and 0.092 for sequence reading against pooling (Figure~\ref{fig:magnitude}). Before a team compares models on maintenance text it should therefore know which record the model will read in service. Capture practice was not manipulated, so 0.456 is the difference between two records as they are written today and says nothing about the return from changing intake.

The two GE gaps have different interpretations, because the customer-to-technician transition introduces new diagnostic evidence whereas the later repair-action field can also restate the completed outcome. A model reading the customer field cannot recover findings that did not exist when the customer wrote, and can at best predict them. Part of the 0.127 gap between the technician and repair-action fields is also restatement: a keyword rule over the curated outcome terms alone reaches 0.292 on that field, and masking those terms costs the BiLSTM 0.038. In the notation of Section~\ref{sec:formulation}, the customer-to-technician transition necessarily changes $H_i(\tau)$ because diagnosis has occurred, but it also changes $g_v$, and $\Delta_R$ does not separate the two. The technician-to-repair transition again changes both, and the later field can also state the completed maintenance action directly. The intervention implied by a limitation in $H_i(\tau)$, capturing more before the decision, differs from the one implied by $g_v$, better models, standardised wording or curation, and the matched-record comparison gives their joint size and not the return from either alone.

The small ASRS synopsis gain of 0.015 arises inside the analyst workflow that also produces the target, so it is best read as evidence about curation within one workflow (Section~\ref{sec:formulation}). Field purpose, by contrast, organises the NHTSA ordering, which holds under masking, out-of-time evaluation and near-duplicate grouping. Across the systems the observed patterns are consistent with differences in producer, workflow stage and record purpose, although those factors are not separately identified.

In the secondary ASRS analyses the lexical baseline and the sequence model also change order between the reporter narrative and the analyst synopsis in every training pair, while RoBERTa and the BiLSTM show no consistent record-dependent difference. A model advantage measured on one record therefore need not transfer unchanged to another record of the same cases.

For the GE triage setting, the scale of the record difference shifts attention from model choice to information availability. The technician record is not thereby a better input for triage, because it is created only after diagnosis. It identifies the information available at triage, where a model reads the customer field and reaches 0.327, as the constraint that no tested model change relieves. Structured intake or access to further contemporaneous records could address that constraint, although the benefit of either intervention has not been measured here. Table~\ref{tab:reporting} illustrates, for one record per system, the context such an evaluation should report: when the record exists, who produced it for what purpose, the decision it would inform and the relation of the label to the records.

\begin{table}[!b]\centering
\caption{Illustrative decision contexts for interpreting record-based classifier evaluations, one record per system. The decision column names the use each evaluation would inform; only the GE decision is a documented operational use. The ASRS label uses information beyond the synopsis; the NHTSA field is one part of the filing the regulator codes from.}\label{tab:reporting}
\footnotesize\setlength{\tabcolsep}{3pt}
\begin{tabular}{@{}llll>{\raggedright\arraybackslash}p{2.1cm}>{\raggedright\arraybackslash}p{2.55cm}@{}}
\toprule
System & Record (producer) & Available at & Record purpose & Illustrative decision & Relation to label \\
\midrule
GE & Technician report (technician) & diagnosis & fault isolation & part prediction at diagnosis & independent: parts transaction \\
ASRS & Analyst synopsis (analyst) & processing & case summary & post-coding review & co-produced with the coding workflow \\
NHTSA & Defect summary (manufacturer) & filing & defect description & component coding & regulator codes from the full filing \\
\bottomrule
\end{tabular}
\end{table}

\begin{figure}[!b]\centering
\includegraphics[width=\textwidth]{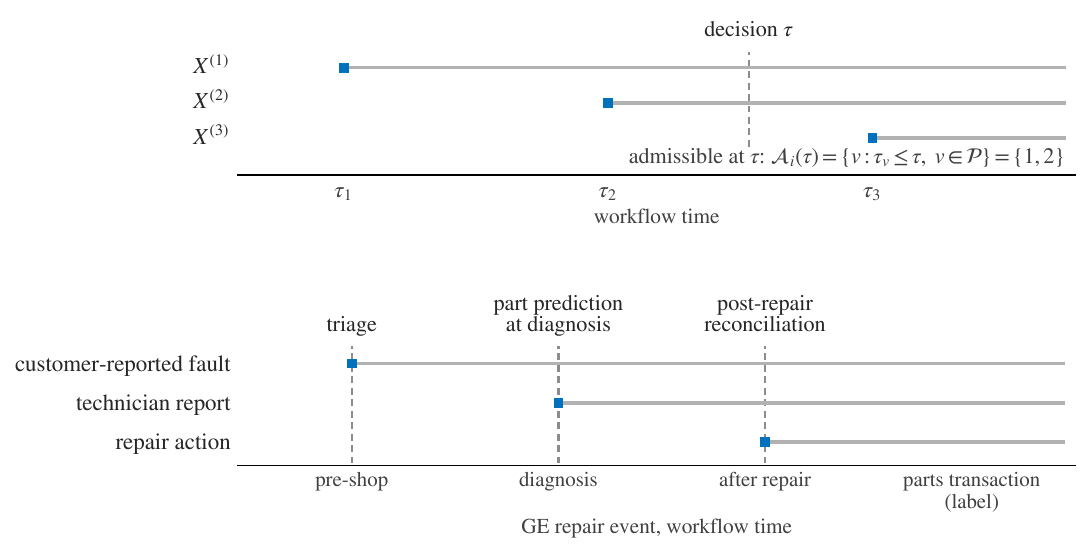}
\caption{Decision-time availability of records. Above, the general case: each record $X^{(v)}$ becomes available at its production time $\tau_v$, and a decision taken at time $\tau$ may read only the records available by then that the deployed system is permitted to access. Below, the GE repair event: the customer-reported fault exists before shop work, the technician report after diagnosis and the repair action after the repair, and the dashed lines mark the decisions of triage, part prediction at diagnosis and post-repair reconciliation. The records whose lines have begun at a decision, and that the deployed system may read, form its admissible set, Equation~\ref{eq:admissible}. The parts transaction, from which the label is taken, follows the repair action.}
\label{fig:decision}
\end{figure}

\enlargethispage{-4\baselineskip}
For retrieval- and tool-using systems, decision-time validity becomes an access-control problem~\cite{Farahani2026,CrespoMarquez2026}: the evaluation has to fix the admissible set of Equation~\ref{eq:admissible}, records that exist by the decision and that the system may read, for the decision assessed (Figure~\ref{fig:decision}). A post-repair record retrieved at run time lies outside the triage set exactly as it would if supplied as a fixed input. In a maintenance system that boundary is the record's creation time against the decision's time, both already stored.

\enlargethispage{-5\baselineskip}
The numerical effects should not be generalised beyond the systems studied. The GE difference bundles a change in what was known ($H_i$) with changes in author, purpose, length and vocabulary ($g_v$), which move together with the field. The ordering holds in every training and under the lexical and recurrent models tested, but the attribution of the 0.456 among those causes stays open. The GE evidence covers one LRU family, one organisation, about a dozen record authors and two assemblies, so rare faults and other sites are untested. The ASRS task is balanced by undersampling and capped at 256 tokens, with a synopsis co-produced with the target, so it does not estimate performance at operational prevalence, and NHTSA covers one regulator and sixteen top-level component classes.

The review evidence is weaker still. The audit covered only misclassified records, unblinded and without an agreement statistic, so its 126 of 168 attributions to narrative--transaction inconsistency do not estimate how common that inconsistency is, and disagreement as a review aid needs blinded, controlled validation before use in screening (Section~S6).

A prospective information-capture intervention would give the clearest causal follow-up: an intake form altered to elicit specific diagnostic detail, with downstream performance measured again at the same decision point and under the same models, would separate the bundled causes of the GE gap. A blinded audit with correctly classified controls would test the review finding, and replication across further LRU families, sites and maintenance systems would show how far the observed sizes carry. The same problem leads to active information acquisition, a system that identifies which observation a decision still lacks and asks for it before the decision is taken.

\section{Conclusion}\label{sec:conclusion}
Which record a classifier reads moved its score on the GE repair events more than any model change we tested. The gap between the customer and technician fields, both written before the parts transaction, was 0.456 macro-F1, against 0.092 for the largest modelling change tested. Smaller and differently shaped dependences appeared in the two public systems, each consistent with who wrote the record, when and for what purpose, and some model comparisons changed with the record. The GE fields are not interchangeable, because each exists at a different workflow stage and supports a different decision. Accordingly, an evaluation should identify the decision being supported, restrict inputs to the records available at that point, document how the target was constructed and compare modelling approaches within that information boundary. The design applies wherever a case carries more than one routine record, one target whose production can be established and a decision point at which the availability of each record is known.

\section*{Data Availability}
ASRS reports are public at \url{https://asrs.arc.nasa.gov/search/database.html} and NHTSA campaigns come from the public recall API~\cite{NHTSAapi}. The analysis code, public-data snapshots, analysis plans and archived public-system results are at \url{https://github.com/ihshaish/source-variation}. The GE Aerospace dataset and the Avi2Vec vectors are proprietary and cannot be shared.

\section*{CRediT author statement}
\textbf{H.I. and P.M.:} Conceptualization, Methodology, Formal analysis, Writing (original draft, review and editing), with Software and Data curation by P.M. and Supervision by H.I. \textbf{T.M.A.Z. and A.D.A.:} Methodology, Validation, Writing (review and editing).

\section*{Funding and conflicts of interest}
This research was supported by GE Aerospace. P.M. and A.D.A.\ are employees of GE Aerospace and contributed in the author roles listed above. Beyond those authorial contributions, the sponsor had no role in study design, analysis, interpretation or the decision to publish.
\bibliographystyle{unsrt}
\bibliography{paper_references}

@inproceedings{Mayhew2023,
  author    = {Mayhew, Peter and Ihshaish, Hisham and Deza, Juan Ignacio and Del Amo, Ana},
  title     = {Maintenance Automation Using Deep Learning Methods: A Case Study from the Aerospace Industry},
  booktitle = {Artificial Neural Networks and Machine Learning -- ICANN 2023},
  series    = {Lecture Notes in Computer Science},
  volume    = {14263},
  pages     = {295--307},
  publisher = {Springer},
  year      = {2023},
  doi       = {10.1007/978-3-031-44204-9_25}
}

@inproceedings{Pennington2014a,
  author    = {Pennington, Jeffrey and Socher, Richard and Manning, Christopher D.},
  title     = {{GloVe}: Global Vectors for Word Representation},
  booktitle = {Proceedings of the 2014 Conference on Empirical Methods in Natural Language Processing (EMNLP)},
  pages     = {1532--1543},
  year      = {2014},
  doi       = {10.3115/v1/D14-1162}
}

@article{Mikolov2013,
  author  = {Mikolov, Tomas and Chen, Kai and Corrado, Greg and Dean, Jeffrey},
  title   = {Efficient Estimation of Word Representations in Vector Space},
  journal = {arXiv preprint arXiv:1301.3781},
  year    = {2013}
}

@inproceedings{Chiu2016,
  author    = {Chiu, Billy and Crichton, Gamal and Korhonen, Anna and Pyysalo, Sampo},
  title     = {How to Train Good Word Embeddings for Biomedical {NLP}},
  booktitle = {Proceedings of the 15th Workshop on Biomedical Natural Language Processing (BioNLP)},
  pages     = {166--174},
  year      = {2016},
  doi       = {10.18653/v1/W16-2922}
}

@article{Nadeau2003,
  author  = {Nadeau, Claude and Bengio, Yoshua},
  title   = {Inference for the Generalization Error},
  journal = {Machine Learning},
  volume  = {52},
  number  = {3},
  pages   = {239--281},
  year    = {2003},
  doi     = {10.1023/A:1024068626366}
}

@article{Tanguy2016,
  author  = {Tanguy, Ludovic and Tulechki, Nikola and Urieli, Assaf and Hermann, Eric and Raynal, C{\'e}line},
  title   = {Natural Language Processing for Aviation Safety Reports: From Classification to Interactive Analysis},
  journal = {Computers in Industry},
  volume  = {78},
  pages   = {80--95},
  year    = {2016},
  doi     = {10.1016/j.compind.2015.09.005}
}

@article{Robinson2015,
  author  = {Robinson, Stephen D. and Irwin, William J. and Kelly, Thomas K. and Wu, Xiaomei O.},
  title   = {Application of Machine Learning to Mapping Primary Causal Factors in Self-Reported Safety Narratives},
  journal = {Safety Science},
  volume  = {75},
  pages   = {118--129},
  year    = {2015},
  doi     = {10.1016/j.ssci.2015.02.003}
}

@inproceedings{MaintNet,
  author    = {Akhbardeh, Farhad and Desell, Travis and Zampieri, Marcos},
  title     = {{MaintNet}: A Collaborative Open-Source Library for Predictive Maintenance Language Resources},
  booktitle = {Proceedings of the 28th International Conference on Computational Linguistics: System Demonstrations},
  pages     = {7--11},
  year      = {2020},
  doi       = {10.18653/v1/2020.coling-demos.2}
}

@inproceedings{Nanyonga,
  author    = {Nanyonga, Aziida and Wasswa, Hassan and Wild, Graham},
  title     = {Phase of Flight Classification in Aviation Safety Using {LSTM}, {GRU}, and {BiLSTM}: A Case Study with {ASN} Dataset},
  booktitle = {2023 International Conference on High Performance Big Data and Intelligent Systems (HDIS)},
  address   = {Macau, China},
  pages     = {24--28},
  year      = {2023},
  doi       = {10.1109/HDIS60872.2023.10499521}
}

@article{Pelt2019,
  author  = {Pelt, Mathijs and Stamoulis, Konstantinos and Apostolidis, Asteris},
  title   = {Data Analytics Case Studies in the Maintenance, Repair and Overhaul ({MRO}) Industry},
  journal = {MATEC Web of Conferences},
  volume  = {304},
  pages   = {04005},
  year    = {2019},
  doi     = {10.1051/matecconf/201930404005}
}

@article{Candell2009,
  author  = {Candell, Olov and Karim, Ramin and S{\"o}derholm, Peter},
  title   = {{eMaintenance}---Information Logistics for Maintenance Support},
  journal = {Robotics and Computer-Integrated Manufacturing},
  volume  = {25},
  number  = {6},
  pages   = {937--944},
  year    = {2009},
  doi     = {10.1016/j.rcim.2009.04.005}
}

@inproceedings{ThurstonB2018,
  author    = {Sexton, Thurston and Hodkiewicz, Melinda and Brundage, Michael P. and Smoker, Thomas},
  title     = {Benchmarking for Keyword Extraction Methodologies in Maintenance Work Orders},
  booktitle = {Proceedings of the Annual Conference of the Prognostics and Health Management Society (PHM)},
  volume    = {10},
  number    = {1},
  year      = {2018},
  doi       = {10.36001/phmconf.2018.v10i1.541}
}

@inproceedings{Chandra2023,
  author    = {Chandra, Chetan and Jing, Xiao and Bendarkar, Mayank V. and Sawant, Kshitij and Elias, Lidya and Kirby, Michelle and Mavris, Dimitri N.},
  title     = {Aviation-{BERT}: A Preliminary Aviation-Specific Natural Language Model},
  booktitle = {AIAA AVIATION 2023 Forum},
  year      = {2023},
  note      = {AIAA 2023-3436},
  doi       = {10.2514/6.2023-3436}
}

@article{TikayatRay2023,
  author  = {Tikayat Ray, Archana and Cole, Bjorn F. and Pinon Fischer, Olivia J. and White, Ryan T. and Mavris, Dimitri N.},
  title   = {aero{BERT}-Classifier: Classification of Aerospace Requirements Using {BERT}},
  journal = {Aerospace},
  year    = {2023},
  volume  = {10},
  number  = {3},
  pages   = {279},
  doi     = {10.3390/aerospace10030279}
}

@article{YangHuang2023,
  author  = {Yang, Chuyang and Huang, Chenyu},
  title   = {Natural Language Processing ({NLP}) in Aviation Safety: Systematic Review of Research and Outlook into the Future},
  journal = {Aerospace},
  year    = {2023},
  volume  = {10},
  number  = {7},
  pages   = {600},
  doi     = {10.3390/aerospace10070600}
}

@inproceedings{Nanyonga2025review,
  author    = {Nanyonga, Aziida and Joiner, Keith and Turhan, Ugur and Wild, Graham},
  title     = {Applications of Natural Language Processing in Aviation Safety: A Review and Qualitative Analysis},
  booktitle = {AIAA SciTech 2025 Forum},
  year      = {2025},
  note      = {AIAA 2025-2153; arXiv:2501.06210},
  doi       = {10.2514/6.2025-2153}
}

@article{Brundage2021,
  author  = {Brundage, Michael P. and Sexton, Thurston and Hodkiewicz, Melinda and Dima, Alden and Lukens, Sarah},
  title   = {Technical Language Processing: Unlocking Maintenance Knowledge},
  journal = {Manufacturing Letters},
  year    = {2021},
  volume  = {27},
  pages   = {42--46},
  doi     = {10.1016/j.mfglet.2020.11.001}
}

@inproceedings{Conte2021,
  author    = {Conte, Anna and Bolland, Coline and Phan, Lynn and Brundage, Michael and Sexton, Thurston},
  title     = {Impact of Data Quality on Maintenance Work Order Analysis: A Case Study in Historical {HVAC} Maintenance Work Orders},
  booktitle = {Proceedings of the PHM Society European Conference},
  year      = {2021},
  volume    = {6},
  number    = {1},
  doi       = {10.36001/phme.2021.v6i1.2814}
}

@article{Kala2022,
  author  = {K{\'a}la, Martin and Lali{\v s}, Andrej and Vojt{\v e}ch, Tom{\'a}{\v s}},
  title   = {Analyzing Aircraft Maintenance Findings with Natural Language Processing},
  journal = {Transportation Research Procedia},
  year    = {2022},
  volume  = {65},
  pages   = {238--245},
  doi     = {10.1016/j.trpro.2022.11.028}
}

@inproceedings{Scott2024,
  author    = {Scott, Michael J. and Kirkpatrick, Oliver and Verhagen, Wim and Kekoc, Vlado and Teunisse, Bob and Zhang, Jenny and Fayek, Haytham and Marzocca, Pier},
  title     = {Application of Natural Language Processing for Aircraft Defect Tracking in Maintenance Operations},
  booktitle = {Proceedings of the 34th Congress of the International Council of the Aeronautical Sciences (ICAS)},
  year      = {2024},
  address   = {Florence, Italy}
}

@inproceedings{Kumar2025,
  author    = {Kumar, Aman and Farahat, Ahmed and Gupta, Chetan},
  title     = {Predicting Maintenance Actions from Historical Logs Using Domain-Specific {LLMs}},
  booktitle = {Proceedings of the Asia Pacific Conference of the PHM Society},
  year      = {2025},
  volume    = {5},
  number    = {1},
  doi       = {10.36001/phmap.2025.v5i1.4652},
  note      = {Published online 13 January 2026}
}

@article{Geirhos2020,
  author  = {Geirhos, Robert and Jacobsen, J{\"o}rn-Henrik and Michaelis, Claudio and Zemel, Richard and Brendel, Wieland and Bethge, Matthias and Wichmann, Felix A.},
  title   = {Shortcut Learning in Deep Neural Networks},
  journal = {Nature Machine Intelligence},
  volume  = {2},
  pages   = {665--673},
  year    = {2020},
  doi     = {10.1038/s42256-020-00257-z}
}

@article{Kapoor2023,
  author  = {Kapoor, Sayash and Narayanan, Arvind},
  title   = {Leakage and the Reproducibility Crisis in Machine-Learning-Based Science},
  journal = {Patterns},
  volume  = {4},
  number  = {9},
  pages   = {100804},
  year    = {2023},
  doi     = {10.1016/j.patter.2023.100804}
}

@article{Deloose2023,
  author  = {Deloose, Arne and Gysels, Glenn and De Baets, Bernard and Verwaeren, Jan},
  title   = {Combining natural language processing and multidimensional classifiers to predict and correct {CMMS} metadata},
  journal = {Computers in Industry},
  volume  = {145},
  pages   = {103830},
  year    = {2023},
  doi     = {10.1016/j.compind.2022.103830}
}

@article{Naqvi2024,
  author  = {Naqvi, Syed Meesam Raza and Ghufran, Mohammad and Varnier, Christophe and Nicod, Jean-Marc and Javed, Kamran and Zerhouni, Noureddine},
  title   = {Unlocking maintenance insights in industrial text through semantic search},
  journal = {Computers in Industry},
  volume  = {157--158},
  pages   = {104083},
  year    = {2024},
  doi     = {10.1016/j.compind.2024.104083}
}

@article{Hershowitz2024,
  author  = {Hershowitz, Brad and Hodkiewicz, Melinda and Bikaun, Tyler and Stewart, Michael and Liu, Wei},
  title   = {Causal knowledge extraction from long text maintenance documents},
  journal = {Computers in Industry},
  volume  = {161},
  pages   = {104110},
  year    = {2024},
  doi     = {10.1016/j.compind.2024.104110}
}

@article{Giordano2024,
  author  = {Giordano, Vito and Fantoni, Gualtiero},
  title   = {Decomposing maintenance actions into sub-tasks using natural language processing: A case study in an {I}talian automotive company},
  journal = {Computers in Industry},
  volume  = {164},
  pages   = {104186},
  year    = {2025},
  doi     = {10.1016/j.compind.2024.104186}
}

@article{LiDataIssues2025,
  author  = {Li, Xuejiao and Cheng, Yang and M{\o}ller, Charles and Lee, Jay},
  title   = {Data issues in industrial {AI} systems: A meta-review and research strategy},
  journal = {Computers in Industry},
  volume  = {173},
  pages   = {104361},
  year    = {2025},
  doi     = {10.1016/j.compind.2025.104361}
}

@inproceedings{Ding2023,
  author    = {Ding, Xiruo and Sheng, Zhecheng and Yeti{\c{s}}gen, Meliha and Pakhomov, Serguei and Cohen, Trevor},
  title     = {Backdoor adjustment of confounding by provenance for robust text classification of multi-institutional clinical notes},
  booktitle = {AMIA Annual Symposium Proceedings},
  year      = {2023},
  pages     = {923--932}
}

@misc{ASRSaboutdata,
  author       = {{NASA Aviation Safety Reporting System}},
  title        = {About {ASRS} data, {ASRS} Database Online},
  howpublished = {\url{https://asrs.arc.nasa.gov/search/dbol/aboutdata.html}},
  note         = {Accessed August 2026}
}

@misc{ASRSprogram,
  author       = {{NASA Aviation Safety Reporting System}},
  title        = {{ASRS} program briefing},
  howpublished = {\url{https://asrs.arc.nasa.gov/docs/ASRS_ProgramBriefing.pdf}},
  note         = {Data through December 2025; accessed August 2026}
}

@misc{NHTSAapi,
  author       = {{National Highway Traffic Safety Administration}},
  title        = {{NHTSA} datasets and {APIs}: recalls},
  howpublished = {\url{https://www.nhtsa.gov/nhtsa-datasets-and-apis}},
  note         = {Accessed August 2026}
}

@article{Yu2024multiview,
  author  = {Yu, Zhiwen and Dong, Ziyang and Yu, Chenchen and Yang, Kaixiang and Fan, Ziwei and Chen, C. L. Philip},
  title   = {A review on multi-view learning},
  journal = {Frontiers of Computer Science},
  year    = {2025},
  volume  = {19},
  number  = {7},
  pages   = {197334},
  doi     = {10.1007/s11704-024-40004-w}
}

@article{Mironczuk2026fusion,
  author  = {Miro{\'n}czuk, Marcin Micha{\l}},
  title   = {Document classification pattern recognition via information fusion: a systematic review of multimodal and multiview representation approaches},
  journal = {Information Fusion},
  volume  = {132},
  pages   = {104247},
  year    = {2026},
  doi     = {10.1016/j.inffus.2026.104247}
}

@article{Mironczuk2019views,
  author  = {Miro{\'n}czuk, Marcin Micha{\l} and Protasiewicz, Jaros{\l}aw and Pedrycz, Witold},
  title   = {Empirical evaluation of feature projection algorithms for multi-view text classification},
  journal = {Expert Systems with Applications},
  volume  = {130},
  pages   = {97--112},
  year    = {2019},
  doi     = {10.1016/j.eswa.2019.04.020}
}

@article{Li2026roberta,
  author  = {Li, Xirui and Romli, Fairuz Izzuddin and Md Ali, Syaril Azrad and Md Zhahir, Md Amzari and Tang, Junqi},
  title   = {Domain-adapted deep learning for aviation incident classification with multiple labels and risk assessment},
  journal = {Engineering Applications of Artificial Intelligence},
  volume  = {173},
  pages   = {114454},
  year    = {2026},
  doi     = {10.1016/j.engappai.2026.114454}
}

@misc{NHTSAcompletion,
  author       = {{National Highway Traffic Safety Administration}},
  title        = {Vehicle safety recall completion rates report: report to {Congress}},
  howpublished = {\url{https://www.nhtsa.gov/sites/nhtsa.dot.gov/files/documents/13376-recall_completion_rates_rtc-tag_final.pdf}},
  note         = {Accessed August 2026}
}

@article{Mikkelsen2026,
  author  = {Mikkelsen, Yngve},
  title   = {Clinical context variables collectively rival model choice in embedding-based retrieval: multi-corpus benchmark study},
  journal = {JMIR Medical Informatics},
  volume  = {14},
  pages   = {e94241},
  year    = {2026},
  doi     = {10.2196/94241}
}

@article{Liu2019roberta,
  author  = {Liu, Yinhan and Ott, Myle and Goyal, Naman and Du, Jingfei and Joshi, Mandar and Chen, Danqi and Levy, Omer and Lewis, Mike and Zettlemoyer, Luke and Stoyanov, Veselin},
  title   = {{RoBERTa}: a robustly optimized {BERT} pretraining approach},
  journal = {arXiv preprint arXiv:1907.11692},
  year    = {2019}
}

@inproceedings{Bean2025construct,
  author    = {Bean, Andrew M. and Kearns, Ryan Othniel and Romanou, Angelika and Hafner, Franziska Sofia and Mayne, Harry and Batzner, Jan and others},
  title     = {Measuring what matters: construct validity in large language model benchmarks},
  booktitle = {Advances in Neural Information Processing Systems (NeurIPS), Datasets and Benchmarks Track},
  year      = {2025},
  note      = {arXiv:2511.04703}
}

@article{Schaffer1988,
  author  = {Schaffer, Simon},
  title   = {Astronomers mark time: discipline and the personal equation},
  journal = {Science in Context},
  volume  = {2},
  number  = {1},
  pages   = {115--145},
  year    = {1988},
  doi     = {10.1017/s026988970000051x}
}

@article{CrespoMarquez2026,
  author  = {Crespo-M{\'a}rquez, Adolfo and G{\'o}mez Fern{\'a}ndez, Juan F.},
  title   = {Agentic {AI} for maintenance management: a process-centric review and a staged framework for industrial adoption},
  journal = {Computers \& Industrial Engineering},
  volume  = {220},
  pages   = {112285},
  year    = {2026},
  doi     = {10.1016/j.cie.2026.112285}
}

@article{Farahani2026,
  author  = {Farahani, Mojtaba A. and Khan, Md Irfan and Wuest, Thorsten},
  title   = {Hybrid agentic {AI} and multi-agent systems in smart manufacturing},
  journal = {Journal of Manufacturing Systems},
  volume  = {86},
  pages   = {612--623},
  year    = {2026},
  doi     = {10.1016/j.jmsy.2026.04.002}
}

@article{Lund2025,
  author  = {Lund, Matthew D.},
  title   = {{``A Vitious Way of Observing'': Kinnebrook and the Prehistory of the Personal Equation}},
  journal = {Isis},
  volume  = {116},
  number  = {3},
  pages   = {461--484},
  year    = {2025},
  doi     = {10.1086/736892}
}

@article{Hochreiter1997,
  author  = {Hochreiter, Sepp and Schmidhuber, J{\"u}rgen},
  title   = {Long short-term memory},
  journal = {Neural Computation},
  volume  = {9},
  number  = {8},
  pages   = {1735--1780},
  year    = {1997},
  doi     = {10.1162/neco.1997.9.8.1735}
}

@article{Schuster1997,
  author  = {Schuster, Mike and Paliwal, Kuldip K.},
  title   = {Bidirectional recurrent neural networks},
  journal = {IEEE Transactions on Signal Processing},
  volume  = {45},
  number  = {11},
  pages   = {2673--2681},
  year    = {1997},
  doi     = {10.1109/78.650093}
}

@article{ProbastAI2025,
  author  = {Moons, Karel G. M. and Damen, Johanna A. A. and Kaul, Tabea and Hooft, Lotty and Andaur Navarro, Constanza L. and Dhiman, Paula and others},
  title   = {{PROBAST+AI}: an updated quality, risk of bias, and applicability assessment tool for prediction models using regression or artificial intelligence methods},
  journal = {BMJ},
  volume  = {388},
  pages   = {e082505},
  year    = {2025},
  doi     = {10.1136/bmj-2024-082505}
}

@inproceedings{Hsu2020notes,
  author    = {Hsu, Chao-Chun and Karnwal, Shantanu and Mullainathan, Sendhil and Obermeyer, Ziad and Tan, Chenhao},
  title     = {Characterizing the value of information in medical notes},
  booktitle = {Findings of the Association for Computational Linguistics: EMNLP 2020},
  pages     = {2062--2072},
  year      = {2020},
  doi       = {10.18653/v1/2020.findings-emnlp.187}
}

@article{Valcamonico2025,
  author  = {Valcamonico, Dario and Baraldi, Piero and Mac\^{e}do, July Bias and Moura, M\'{a}rcio das Chagas and Brown, Jonathan and Gauthier, St\'{e}phane and Zio, Enrico},
  title   = {A systematic procedure for the analysis of maintenance reports based on a taxonomy and {BERT} attention mechanism},
  journal = {Reliability Engineering \& System Safety},
  volume  = {257},
  pages   = {110834},
  year    = {2025},
  doi     = {10.1016/j.ress.2025.110834}
}

@article{Bhardwaj2025,
  author  = {Bhardwaj, Abhijeet Sandeep and Veeramani, Dharmaraj},
  title   = {An unsupervised multi-class ensemble classifier for identifying equipment failure mechanisms from maintenance records},
  journal = {Reliability Engineering \& System Safety},
  volume  = {264},
  pages   = {111410},
  year    = {2025},
  doi     = {10.1016/j.ress.2025.111410}
}

@article{Valcamonico2024,
  author  = {Valcamonico, Dario and Baraldi, Piero and Zio, Enrico and Decarli, Luca and Crivellari, Anna and La Rosa, Laura},
  title   = {Combining natural language processing and {B}ayesian networks for the probabilistic estimation of the severity of process safety events in hydrocarbon production assets},
  journal = {Reliability Engineering \& System Safety},
  volume  = {241},
  pages   = {109638},
  year    = {2024},
  doi     = {10.1016/j.ress.2023.109638}
}

@article{KarimiDehkordi2025,
  author  = {Karimi Dehkordi, Mohammadreza and Sattari, Fereshteh and Lefsrud, Lianne},
  title   = {Creating an incident investigation framework for a complex socio-technical system: application of multi-label text classification and {B}ayesian network structure learning},
  journal = {Reliability Engineering \& System Safety},
  volume  = {260},
  pages   = {110971},
  year    = {2025},
  doi     = {10.1016/j.ress.2025.110971}
}

@article{Macedo2026,
  author  = {Macedo, July B. and Ramos, Pl\'{i}nio M. S. and Maior, Caio B. S. and Moura, M\'{a}rcio J. C. and Lins, Isis D.},
  title   = {Automated identification of pilot failures in aviation accidents using a {BERT}-based classifier and topic modeling},
  journal = {Reliability Engineering \& System Safety},
  volume  = {273},
  pages   = {112284},
  year    = {2026},
  doi     = {10.1016/j.ress.2026.112284}
}

@article{Bhardwaj2023,
  author  = {Bhardwaj, Abhijeet Sandeep and Veeramani, Dharmaraj and Zhou, Shiyu},
  title   = {Identifying equipment health status from maintenance records using lexicon based unsupervised sentiment analysis adjusted for negation ({LUSAA-N})},
  journal = {Computers \& Industrial Engineering},
  volume  = {186},
  pages   = {109693},
  year    = {2023},
  doi     = {10.1016/j.cie.2023.109693}
}

@article{Chung2023,
  author  = {Chung, Euisuk and Park, Kyoungchan and Kang, Pilsung},
  title   = {Fault classification and timing prediction based on shipment inspection data and maintenance reports for semiconductor manufacturing equipment},
  journal = {Computers \& Industrial Engineering},
  volume  = {176},
  pages   = {108972},
  year    = {2023},
  doi     = {10.1016/j.cie.2022.108972}
}

@article{LiMeijing2025,
  author  = {Li, Meijing and Li, Ci and Huang, Runqing and Gu, Zixing and Wang, Jinghao and Wang, Yichen and Fan, Jiabin and Zhao, Ziqi and Wang, Yu},
  title   = {{LK-VTC}: a label knowledge-enhanced variational text classification method for port equipment fault diagnosis},
  journal = {Computers \& Industrial Engineering},
  volume  = {210},
  pages   = {111486},
  year    = {2025},
  doi     = {10.1016/j.cie.2025.111486}
}
\clearpage
\includepdf[pages=-]{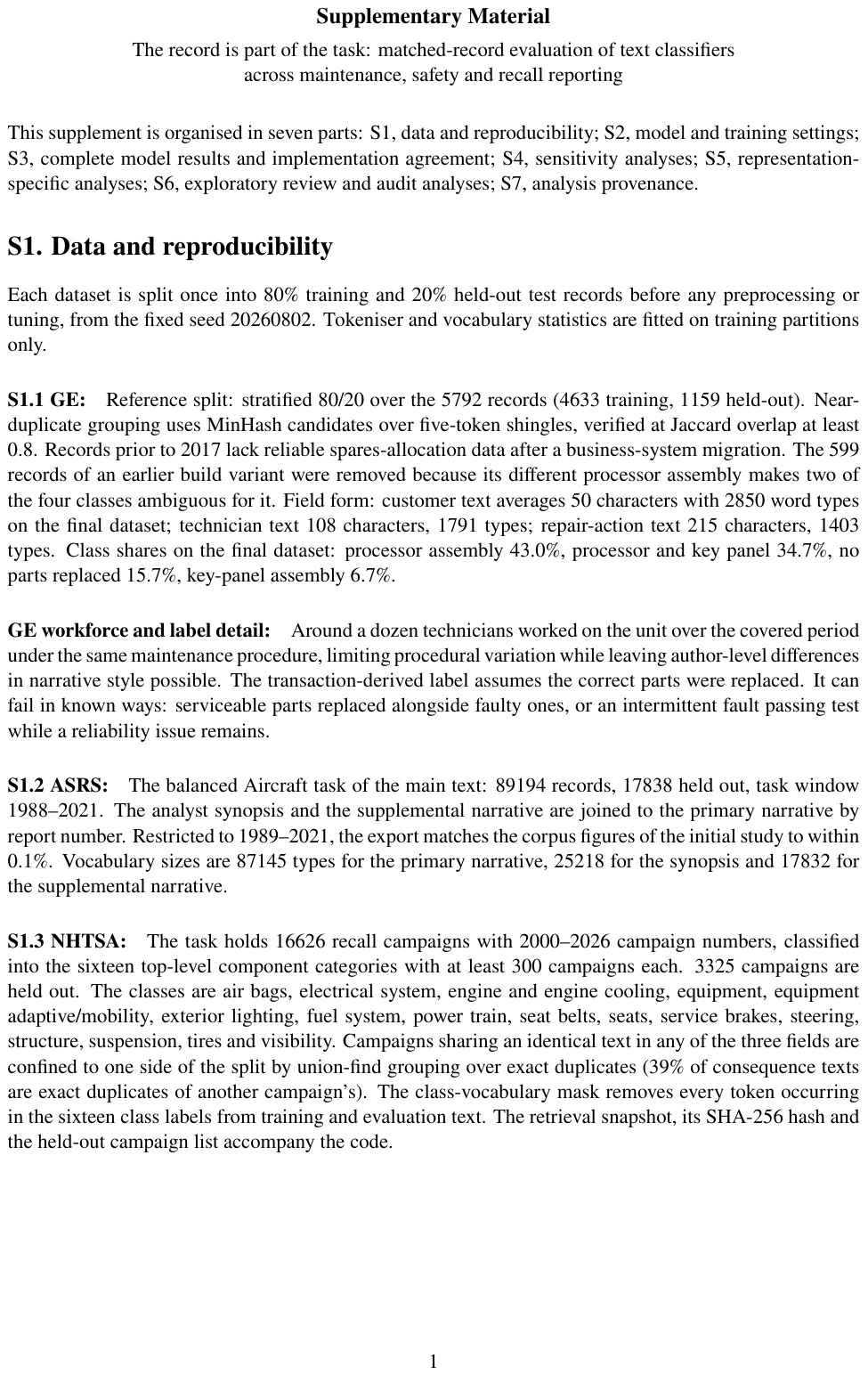}

\end{document}